\documentclass[runningheads]{llncs}

\usepackage[T1]{fontenc}    %
\usepackage{hyperref}       %
\usepackage{url}            %
\usepackage{booktabs}       %
\usepackage{amsfonts}       %
\usepackage{nicefrac}       %
\usepackage{microtype}      %

\usepackage{amsmath}
\usepackage{mathtools}
\usepackage{bm}

\usepackage{subcaption} %

\usepackage{float} 
\usepackage[pdftex]{graphicx}
\usepackage{pdfpages}
\usepackage{wrapfig}

\usepackage{pdfsync,amssymb,ifpdf,color}

\newcommand{\W}[1]{\mathbf{W}^\text{#1}} %
\newcommand{\U}[1]{\mathbf{U}^\text{#1}}

\newcommand{\x}{\mathbf{x}}

\newcommand{\y}{\mathbf{y}} 
\def\r{\mathbf{r}} %
\newcommand{\z}{\mathbf{z}} 
\newcommand{\h}{\mathbf{h}} 

\usepackage{amsmath}

\begin{document}
\sloppy

\title{Task-Synchronized Recurrent Neural Networks}

\author{%
  Mantas Luko\v{s}evi\v{c}ius\orcidID{0000-0001-7963-285X} \and
  Arnas Uselis 
}

\institute{
  Kaunas University of Technology, LT-44249 Kaunas, Lithuania 
  \email{mantas.lukosevicius@ktu.lt} and \email{auselis@gmx.com}
}

\maketitle

\begin{abstract}

Data are often sampled irregularly in time. Dealing with this using Recurrent Neural Networks (RNNs) traditionally involved ignoring the fact, feeding the time differences as additional inputs, or resampling the data. All these methods have their shortcomings. We propose an elegant straightforward alternative approach where instead the RNN is in effect resampled in time to match the time of the data or the task at hand. We use Echo State Network (ESN) and Gated Recurrent Unit (GRU) as the basis for our solution. Such RNNs can be seen as discretizations of continuous-time dynamical systems, which gives a solid theoretical ground to our approach. Our Task-Synchronized ESN (TSESN) and GRU (TSGRU) models allow for a direct model time setting and require no additional training, parameter tuning, or computation (solving differential equations or interpolating data) compared to their regular counterparts, thus retaining their original efficiency. We confirm empirically that our models can effectively compensate for the time-non-uniformity of the data and demonstrate that they compare favorably to data resampling, classical RNN methods, and alternative RNN models proposed to deal with time irregularities on several real-world nonuniform-time datasets. We open-source the code at \url{https://github.com/oshapio/task-synchronized-RNNs}.

\end{abstract}

\section{Introduction}

A usual assumption about a time series is that it is uniformly sampled in time, whether applying Recurrent Neural Networks (RNNs) or other models. In reality, however, this is often not the case for multiple reasons: data is gathered at irregular intervals, like patient examinations or expeditions; data comes from sources that get dated, like archaeological or geological samples, historical written sources; the data is recorded when a certain event happens, like an action by a user, an accident, a natural phenomenon, economic transaction; neural spike signals, etc. This can either be an offline or an online setup requiring a real-time response. Here we deal with the case when the time irregularities are known, i.e., data samples have time stamps. Missing values in time series can also be treated as time irregularity, as long as the data are missing in all the input dimensions at the same time. 

When data is irregularly sampled in time, typically RNNs are not applied, the time irregularity is ignored, the data are resampled by interpolation, or the time irregularities are simply fed as additional inputs in hopes that the model will learn to properly deal with them. All these approaches have their limitations. 

Several specialized RNN architectures have been proposed to address this problem. They typically treat time irregularities as a special kind of input aiming in learning to accommodate it better. We review them and other alternatives in Section \ref{literature}.

Here we take a bit different approach. Many types of RNNs can be seen as a discretization of a continuous-time dynamical system. Based on this, we can re-discretize it online with variable time steps that match the ones in the data (or the task) at hand. This way, since time is relative, we make the RNN ``live'' in the irregular time of the task, as opposed to resampling the data to match the regular time of the RNN. We investigate this in two instances of RNNs: echo state networks and gated recurrent units.

The rest of the article is structured in the following way. We discuss related work in Section \ref{literature} and analytically derive our models in Section \ref{methods}. We then test our approach on two synthetic chaotic high-precision tasks to confirm empirically our analytic derivation in Section \ref{synthetic} and on two real-world datasets: classification of signals with missing values in Section \ref{gesture_data} and prediction of inherently irregular-time series in Section \ref{cave_data}. Finally, we discuss the results, limitations, and possible future work in Section \ref{dicussion}.

An early version of this article titled ``Time-Adaptive Recurrent Neural Networks'' has been submitted and double-blind reviewed for a prominent ML conference in 2019 where it was not accepted. We made the preprint available in 2022 at \url{https://arxiv.org/abs/2204.05192v1}
because of another work \cite{Schake22mthesis} that built on it. We have changed the name of the article and the methods in this latest incarnation because the new name is arguably more precise, and because an article with a very similar title \cite{Kag_2021_CVPR} has been published in the meantime.

\section{Related Work}\label{literature}

Here we give a short overview of more or less related previous approaches dealing with time-irregular sequences.

\subsection{Direct Time to Output Mapping}

One effective, but rather niche class of models capable of dealing with irregularly-sampled data are models that take time as a real-valued input and map it directly to the desired outputs. These feed-forward methods have advantages in that they are stable and can predict not only at the future (forecast) but also at intermediate time points (impute). On the downside, they typically can not be reused for other tasks, must be trained separately on each time series, and do not deal well with multivariate data, and typically can not predict quite far. Any regression model can in principle be used for that but some are better suited than others. Gaussian process regression \cite{NIPS1995_7cce53cf} is a powerful probabilistic example of such. 

Various decompositions of the continuous-time series can also be applied, for example, \cite{GodfreyGashler18}  bridges the gap between Fourier transformation and neural networks. 

In this work, we only consider RNNs that have intrinsic memory and state and process input sequentially.

\subsection{RNN Models with Data Pre-Treatment Layers}

Much of related previous work deals with medical data that are often both irregularly sampled in time and have missing values in some of the input dimensions.

There is a class of methods where the data is ``sanitized'' in the first layer and fed to RNNs above. The missing medical examination values were imputed and the data were resampled regularly in time before feeding them to a Long Short-Term Memory (LSTM) network in \cite{lipton2015learning}. The resampling is prone to information loss. Similar data has been interpolated by a special trainable semi-parametric network, after which, other types of models can be applied in \cite{shukla2018interpolation}. A statistical approach has been taken in \cite{futoma2017learning}, where a Gaussian process is responsible for handling the missing values and time irregularities of the time series, and then passes this information to an RNN. These are more sophisticated methods, but they can suffer from similar information loss since the upper layers receive only the interpolated data. In addition, the computational complexity of Gaussian-process-based methods is high. An RNN attention mechanism adapted to handle multiple periods in data was applied after imputing missing values in \cite{cinar2018period}.

\subsection{RNN Models with Special Gates}

It is also possible to give the time (or its change) simply as one of the regular inputs to the network, hoping that the model will learn to treat it correctly. We also consider, do experiments, and offer some analysis of this option here. The time input can also be encoded into a vector of useful features before feeding it to a model, capturing, e.g., relevant periodicity, like time of day, week, or year; similar to positional encodings in attention-based models.

Several RNN architectures have been introduced that treat time (differences) as a special kind of input for learning.

The authors of \cite{mozer2017discrete} have added special time-scale-dependent states to Gated Recurrent Unit (GRU) network but have concluded, that such an approach is not more effective than a simple GRU with time differences as part of the regular input. Short-term and long-term modeling has been introduced in LSTSMs for giving recommendations from temporal usage data in \cite{ijcai2017-504}. Similarly, GRU-D, a variation of GRU with additional decay units to its gates and inputs based on the missingness of the variables was introduced in \cite{che2018recurrent}. It was observed, that the fact of missing data might also be informative.
A slightly different approach has been taken in \cite{DBLP:journals/corr/PhamTPV16}, where instead of decaying the state, the forgetting of a network was directly influenced by the time lapse between the events. This direct transformation of the forget gate has an impact on global memory which in turn can lead the model to forget about global history based on a longer time gap, which might not be wanted. 

To address this problem, Time-Aware LSTM \cite{BaytasEtAl17} was introduced, which decomposes LSTM's memory cell into two blocks: short-term and long-term memory. The short-term memory reacts to time irregularities in the same way as in previous models, while the long-term memory learns how much information to choose from the short-term memory.

Phased LSTMs \cite{Neil:2016:PLA:3157382.3157532} allow LSTMs to operate on multiple timescales by introducing additional time gates to the LSTM units and making them specialize by opening the gates with different frequencies and phases. The model can also deal with time-irregular data.

These model variations are a bit similar to our approach, however, they are based on experimental gate design and require extra learning for the extra time inputs. Our approach to task-synchronized RNNs is different in that timing is introduced systematically and directly based on time-sampling the RNNs as continuous-time dynamical systems, and requires no additional learning. The previous work that our approaches are directly based on is cited in Section \ref{methods}.

\subsection{RNN Methods Employing Differential Equation Solvers}

Another class of RNN methods, similar to our approach, derives from continuous-time RNNs \cite{FUNAHASHI1993801} that are defined using differential equations.%
 
The irregular time sampling problem has been addressed in Neural Ordinary Differential Equations (Neural ODEs) \cite{chen2018neural}. They enable computing a state of a continuous-time RNN at any desired time point using an ODE solver. The ODE solver, however, does not accommodate RNN inputs, it can only compute the evolution of its autonomous state.

ODE-RNNs proposed in \cite{NEURIPS2019_42a6845a} enable inputs and outputs by mixing the internal dynamics of a continuous-time RNN computed by Neural ODEs in the absence of input with a regular discrete-time RNN update whenever an input arrives. These two types of dynamics, however, do not mix well and are not a very natural way to introduce inputs, not the way they are introduced in classical continuous-time RNNs \cite{FUNAHASHI1993801}. 

For that, the inputs must affect the continuous RNN dynamics directly and continually. This can be done using a Controlled Differential Equation (CDE) solver as presented in \cite{NEURIPS2020_4a5876b4}. For this, however, the irregularly-sampled input is interpolated into a continuous-time signal using a cubic spline. This carries some of the drawbacks of data resampling and somewhat obsoletes the need for irregular-time modeling of the RNN unless the outputs also need to be irregularly sampled in time. 

The latter method is probably the closest conceptually and functionally to our approach, except that we do not interpolate the input data, nor use the differential equation solvers. It can, however, be employed as an alternative to our approach to achieving similar goals, and has in fact been done so in \cite{Schake22mthesis}, which can be seen as a preliminary offshoot of this work.

A model whose state is governed by a linear stochastic differential equation and encoder-decoder setup is used to feed the irregularly-sampled inputs and read the outputs was proposed in \cite{pmlr-v162-schirmer22a}.

Here, due to space limitations, we only went through some of the most prominent and related previous works. A much more comprehensive overview of RNN approaches handling irregular time data can be found in \cite{WEERAKODY2021161} or \cite{Schake22mthesis}.

\section{Methods}\label{methods}

Here we derive our task-synchronized RNNs for two popular types of networks, discuss their differences and similarities, variations.

\subsection{Task-Synchronized Echo State Networks}\label{esn}

Echo State Networks (ESNs) \cite{Jaeger01a} dynamics can be seen as a time-discretization of the differential equation \cite{JaegerEtAl07si}
\begin{equation}
	\dot{\h} = \frac{1}{c}\left(-\alpha \h + \sigma(\W{h} \x + \U{h}\h)\right) \label{eq:dot_esn}, 
\end{equation}
where $\h$ is the internal activation state and $\x$ is the input vector, $\sigma(\cdot)$ is the activation function (usually $\tanh(\cdot)$)%
, ${\W{}}^\cdot$ and ${\U{}}^\cdot$ denote the corresponding input and update weight matrices, $c$ denotes the global time constant and $\alpha$ the leaking rate. Here, and in other equations, bias weights are subsumed in ${\W{}}^\cdot$, assuming that a constant $\mathbf{1}$ is appended to $\x$. Applying linear Euler discretization
\begin{equation}
\dot{\h} \approx \frac{\h_{n} - \h_{n-1}}{\Delta{t}}
\label{eq:euler}
\end{equation}
to \eqref{eq:dot_esn} we get the discrete-time $n$ ESN \cite{JaegerEtAl07si}
\begin{equation}
	\h_n = \left(1 - \alpha \frac{\Delta t}{c}  \right)\h_{n-1}+ \frac{\Delta t}{c} \sigma\left( \W{h} \x_n + \U{h}\h_{n-1} \right).
	\label{eq:discretized_esn}
\end{equation}
Here we take $\alpha \equiv 1$ in \eqref{eq:discretized_esn} to have one meta-parameter less and redefine the leaking rate as $\alpha \equiv \frac{1}{c}$.\footnote{This is a trade-off between simplicity and some performance, which is not necessary to make.} Normally the discretization step $\Delta t$ is constant, and subsuming it in $\alpha = \frac{\Delta t}{c}$ we get the \textbf{typical ESN} update equation 
\begin{equation}
	\h_n = (1 - \alpha)\h_{n-1}+ \alpha \sigma\left(\W{h} \x_n + \U{h}\h_{n-1}\right).
	\label{eq:esn}
\end{equation}

This is not something new so far. For example, a similar derivation is presented in \cite{LukoseviciusEtal06} and \cite{JaegerEtAl07si}, and later in \cite{TallecOllivier18}, not referncing the prior two.

For our \textbf{Task-Synchronized ESN} (TSESN), we allow $\Delta t$ in \eqref{eq:discretized_esn} to be variable in time and set directly from data, yielding
\begin{equation}
	\h_n = (1 - \alpha\Delta t_n)\h_{n-1}+ \alpha\Delta t_n \sigma\left(\W{h} \x_n + \U{h}\h_{n-1}\right).
	\label{eq:taesn}
\end{equation}
The time steps of this model can be adapted to the irregular time steps $\Delta t_n$ of the data, in effect time-resampling the RNN instead of the data.

Readouts $\y_n$ from ESNs are typically done in a linear fashion
\begin{equation}
    \y_n = \W{y}[\x_n;\h_n].
    \label{eq:esn_out}
\end{equation}

\subsection{Task-Synchronized Gated Recurrent Units}\label{gru}

Following the same notation, Gated Recurrent Unit (GRU) \cite{ChoEtal14} networks are governed by
\begin{align}
	\z_{n}&=\sigma_{g}(\W{z}\x_{n}+\U{z}\h_{n-1}),\label{eq:gru_z}\\
    \r_{n}&=\sigma_{g}(\W{r}\x_{n}+\U{r}\h_{n-1}),\label{eq:gru_r}\\
    \h_{n}&=(1-\z_{n})\circ \h_{n-1}+\z_{n}\circ \sigma\left(\W{h}\x_{n}+\U{h}(\r_{n}\circ \h_{n-1})\right),
	\label{eq:gru}
\end{align}
where $\z_t$ is the update and $\r_t$ the reset (or forget) gate vectors, $\sigma_{g}(\cdot)\in (0,1)$ typically stands for a logistic sigmoid, and $\cdot\circ\cdot$ for element-wise multiplication. 

We can observe the similarity between the leaky-integration in ESN \eqref{eq:esn} and gating in GRU \eqref{eq:gru}, which was already noted by the authors of \cite{ChoEtal14}. We can further observe that if GRU \eqref{eq:gru} receives the variable time step $\Delta t_n$ as part of input $\x_n$, %
it can become very similar to TSESN \eqref{eq:taesn} if the reset gate is not used ($\r_n\equiv\mathbf{1}$) and the update gate $\z_n$ is learned to be $\alpha\Delta t_n$, and $\Delta t_n$ is learned to be ignored elsewhere. This somewhat justifies such an approach (providing $\Delta t_n$ as part of input $\x_n$ to GRU) and could explain why it was hard to beat it in \cite{mozer2017discrete}. We refer to this approach as GRU+$\Delta t$ in our experiments.

GRU, however, can in itself be seen as a time-discretization of a continuous-time differential equation %
similar to \eqref{eq:dot_esn}
\begin{equation}
	\dot{\h} = \z\circ\left(-\h + \sigma(\W{h} \x + \U{h}(\r\circ\h))\right) \label{eq:dot_gru}. 
\end{equation}
Applying Euler discretization \eqref{eq:euler} with a constant $\Delta t$ to \eqref{eq:dot_gru}, we get the standard GRU \eqref{eq:gru}. We omit the constants $\frac{1}{c}$ and $\alpha$ here, as well as $\Delta t$, assuming that they can be subsumed in the learned $\z$ which, fittingly, governs the update rate of $\h$ in \eqref{eq:dot_gru}.\footnote{This is again a trade-off, and having more hyper-parameters might somewhat improve the performance.}

Letting the $\Delta t$ remain adaptive in the discretization process of \eqref{eq:dot_gru} by \eqref{eq:euler}, we get \textbf{Task-Synchronized GRU} (TSGRU)
\begin{equation}
    \h_{n}=(1-\Delta t_n\z_{n})\circ \h_{n-1}+(\Delta t_n\z_{n})\circ \sigma\left(\W{h}\x_{n}+\U{h}(\r_{n}\circ \h_{n-1})\right),
	\label{eq:tagru}
\end{equation}
instead of \eqref{eq:gru}, similar to \eqref{eq:taesn}. The update \eqref{eq:gru_z} and reset \eqref{eq:gru_r} gates remain unaffected.

This method, compared to feeding $\Delta t_n$ as part of input $\x_n$ in GRU \eqref{eq:gru}, has no additional trained parameters in ${\W{}}^\cdot$ and no need to learn the role of $\Delta t_n$, freeing the gating mechanism to learn other data-related things.

We use readouts $\y_t$ from all types of GRU similar to ESN \eqref{eq:esn_out}
\begin{equation}
    \y_t = \W{y}[1;\h_t].
    \label{eq:gru_out}
\end{equation}

Despite similarities, ESN and GRU networks are trained very differently. In ESNs only $\W{y}$ \eqref{eq:esn_out} is learned in a one-shot linear regression manner, and the rest of weights remain generated randomly based on a couple of meta-parameters \cite{Lukosevicius12a}. GRU networks, conversely, are fully end-to-end trained using error back-propagation and gradient descent \cite{ChoEtal14}. This makes GRUs more expressive at a cost of training time. 

Note, that the model we define in Section \ref{esn} is in fact a classical RNN with leaky-integrator units, that could also be fully trained using gradient methods.

\subsection{Nonlinear Time Scaling}\label{fdt}

Having big $\Delta t_n$ values is a problem for our model, as $\alpha\Delta t_n$ in \eqref{eq:taesn} and $\Delta t_n\z_n$ in \eqref{eq:tagru} should be $\leq 1$. We scale $\Delta t_n$ to $[0,1]$ dividing them by the maximum, but with large outliers this might result in many minuscule values. 

For these practical reasons we also investigate a replacement of $\Delta t_n$ with its nonlinear function $f(\Delta t_n)$ in \eqref{eq:taesn}, \eqref{eq:tagru}, and also where $\Delta t_n$ comes as additional input in other models. This amounts to redefining the derivatives \eqref{eq:dot_esn} and \eqref{eq:dot_gru} as being not with respect to $d t$ but $f(d t)$. In particular we consider $f(\Delta t_n)=1-e^{-\Delta t_n}$. Models having this replacement we denote by ``exp'' in our experiments.

An alternative way to deal with large $\Delta t_n$ would be to interpolate/infill the data where the gaps are too wide.

Note that our task-synchronized versions of RNNs are generalizing extensions that fall back to regular versions when $\Delta t_n$ is constant, provided that it is normalized.

\section{Experiments}\label{experiments}

\subsection{Synthetic Chaotic Attractor Datasets}\label{synthetic}

To empirically test the validity of our approach we first turn to high-precision synthetic tasks. ESNs are known for their state-of-the-art performance in predicting some types of chaotic attractors \cite{JaegerHaas04}. This high precision comes in part from applying linear regression instead of stochastic gradient descent. 

We artificially introduce increasing time irregularities to such data and investigate if TSESN can compensate for them and maintain the state-of-the-art performance. ESN+$\Delta t$, a regular ESN receiving $\Delta t_n$ as additional input is also added for comparison. 

In the first experiment we use Lorenz chaotic attractor \cite{Lorenz63} with parameters $\sigma = 10, \beta = \frac{8}{3}$ and $\rho = 28$. We introduce time irregularity factor $\pi$ and generate the 3D Lorenz attractor data with $\Delta t_n$ uniformly sampled from the $(\mathrm{max}(0,0.01 - \pi), 0.01 + \pi]$. The time is regular at $\Delta t_n\equiv 0.01$ with $\pi=0$ and becomes more irregular with increasing $\pi$.

We first generate 10\,000 timesteps with $\pi=0$ and split the data into 60\% training, 20\% validation, and 20\% testing sets. We use (TS)ESNs with 500 internal units and grid-search their $\alpha$, spectral radius of $\U{h}$, and regularization of $\W{y}$ to find the best meta-parameters by training each model and comparing its generated sequences with validation. Note that we only select the hyperparameters for the ESN model; identical hyperparameters are then used for the ESN+$\Delta t$ and TSESN.

\begin{figure}[H]	
	  \begin{subfigure}[t]{0.5\textwidth}
        \centering
        \caption{Lorenz} 
        \includegraphics[width=\textwidth]{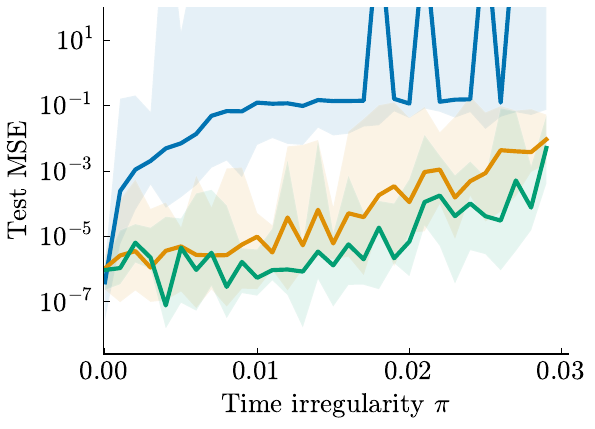}
       \label{fig:lorenz}%
    \end{subfigure}%
	\begin{subfigure}[t]{0.5\textwidth}%
        \centering%
        \caption{Mackey-Glass} 
        \includegraphics[width=\textwidth]{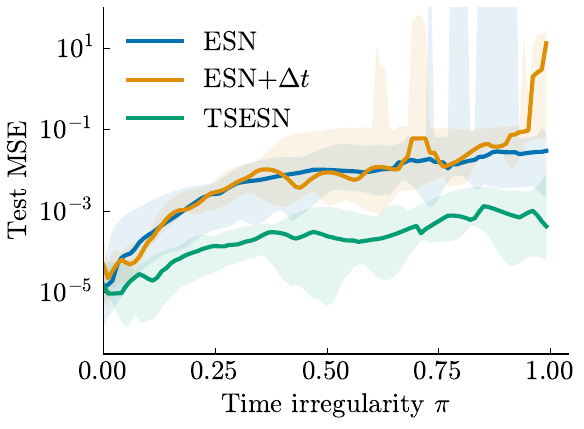}
        \label{fig:mackey-glass}
    \end{subfigure}
    \vspace{-2.em}
    \caption{Comparison of ESN, ESN+$\Delta t$, and TSESN mean error for varying time irregularity of chaotic attractor data. Shading indicates min/max errors over 50 runs.}
    \label{fig:chaotic}
\end{figure}

Having the good ESN and TSESN models (which are initially identical) we then test them with ever more time-irregular data, increasing $\pi$ each time by 0.001, and each time initializing and testing the models on 50 200-step generative runs. The performances over the time-irregularity $\pi$, which goes to high extremes, are presented in Figure \ref{fig:lorenz}.
\vspace{-1.5em}
\vspace{-1em}

We see that the performance of TSESN degrades very slowly with increasing time irregularity, which confirms the validity of our approach, while the identical regular ESN becomes unstable (confused) quite fast when encountering time-irregular data. Curiously, ESN+$\Delta t$ is not far behind TSESN in this task.

We also did a similar test on a more memory-demanding Mackey-Glass \cite{MackeyGlass77} chaotic attractor with parameters $\beta = 0.2$, $n = 10$, $\gamma = 0.1$, and $\tau = 17$. Since the attractor depends on a fixed time interval $\tau$, we had to first generate the data with a small (for this dataset) uniform $\Delta t'=0.01$, and then produce the time-irregular data by cubic spline interpolation from this high-regular-sampling-rate data, uniformly sampling $\Delta t_n$ from $(1-\pi,1+\pi]$.

The results are presented in Figure \ref{fig:mackey-glass}. We can see that TSESN is again much more robust to time irregularity, but ESN also degrades less in this task. Curiously again, ESN+$\Delta t$ is not better than regular ESN in this task and for high $\pi$ is even worse.

The GRU and LSTM RNN models performed expectedly poorly on these high-precision clean synthetic tasks because they use noisy stochastic gradient descent to learn their weights.

\subsection{UWave Gesture Dataset}\label{gesture_data}
UWaveGestureAll \cite{liu2009uwave}\footnote{UWaveGestureLibraryAll is available at \url{http://timeseriesclassification.com}.} is a univariate time series data set consisting of gesture patterns of eight types. The dataset contains 890 training  and 3\,580 testing samples, where each sample consists of 900 data points. We follow the methodology used in \cite{Futoma2017} and \cite{shukla2018interpolation} by sampling random 10\% out of every sample. For evaluation, 30\% of the training data was used for validation. Each trainable model was iterated for 100 epochs for 10 times with random models' initialization. Also, each ESN variation had 500 neurons in its reservoir, while the other baselines had 100 neurons in their hidden layers. Moreover, the leaking rate and spectral radius scaling hyper-parameters for ESN variations were found using grid search.

\begin{figure}
    \centering
    \includegraphics{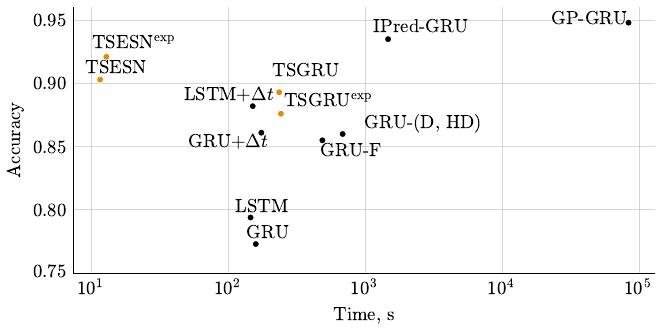}
    \caption{ Time and performance comparison on UWaveGesture dataset. Proposed variations are marked in orange. Simple ESN was omitted due to subpar performance on this task. }  %
    \label{fig:keys_count}
\end{figure}
\vspace{-1em}

Furthermore, we compare the results reported in \cite{shukla2018interpolation}. Concretely, we report the following baselines, that performed most notably: GRU-F, where the missing values are imputed with the last observation of the series, GRU-D \cite{che2018recurrent}, which adds exponential decaying in the input and the hidden state, when the variable is not observed, GRU-HD \cite{che2018recurrent}, where the decay is only introduced in the hidden state and not the input, GP-GRU \cite{futoma2017learning}, a Gaussian process, with GRU as a classifier, and IPred-GRU \cite{shukla2018interpolation}, an interpolation-prediction network with GRU as a classifier. Each of the baselines had 128 neurons in GRU classifier hidden state. The training time was scaled to match the hardware differences with \cite{shukla2018interpolation}.

From each of the training sessions, the best model based on the validation loss was chosen. Averaged results are presented in Table \ref{tab:res_uwave}.

\setlength{\tabcolsep}{8pt}
\begin{table}[htb]
    \centering
    \caption{Results on UWaveGestureAll dataset. *results reported by \cite{shukla2018interpolation}.}
    \begin{tabular}{ l r l r r}
    \toprule
     Model  & Parameters & Accuracy & Training time, s & $f(\Delta t_n)$\\
        \midrule
GRU-F* & 50,952 & 0.855 & $4.840 \times 10^2$ & - \\
GRU-D* \cite{che2018recurrent} & ? & 0.860 &$6.810 \times 10^2$ & exp \\
GRU-HD* \cite{che2018recurrent} & ? & 0.860 & $6.900 \times 10^2$ & exp \\
GP-GRU* \cite{futoma2017learning} & ? & 0.948 & $8.365 \times 10^4$ & - \\
IPred-GRU*  \cite{shukla2018interpolation} & 51,721 & 0.935 & $1.463 \times 10^3$ & - \\
\midrule

ESN & 4,016 & 0.567 $\pm$ 0.265& $1.101 \times 10^1$  & linear \\
\textbf{TSESN} & 4,016 & 0.921 $\pm$ 0.006 & $1.277 \times 10^1$ & exp \\
\textbf{TSESN}& 4,016 & 0.903 $\pm$ 0.011 & $1.150 \times 10^1$ & linear  \\
\midrule
LSTM  & 41,608 & 0.794 $\pm$ 0.016& $1.445 \times 10^2 $ & -\\
LSTM+$\Delta t$ & 42,008 & 0.882 $\pm$ 0.009  & $1.500 \times 10^2 $ & linear \\
GRU & 31,408 & 0.773 $\pm$ 0.009& $1.579 \times 10^2 $ & -\\
GRU+$\Delta t$ & 31,708  &0.861 $\pm$ 0.007 & $1.733 \times 10^2 $ & linear\\
\textbf{TSGRU} & 31,408 & 0.893 $\pm$ 0.010  &$2.341 \times 10^2 $ & linear \\
\textbf{TSGRU} & 31,408 & 0.876 $\pm$ 0.012& $2.410 \times 10^2 $ & exp \\
\bottomrule
    \end{tabular}
    \label{tab:res_uwave}        
\end{table}

From the results in Table \ref{tab:res_uwave} we can see that TSESNs proved to be very good in both performance and training time. We found that the regular ESN was quite sensitive to its random weight generation in this time-irregular task, which explains its poor performance and high standard deviation. TSGRU also performed significantly better than regular GRU or LSTM models and slightly better compared to versions GRU+$\Delta t$ and LSTM+$\Delta t$ that receive $\Delta t_n$ as inputs. The training time of TSGRU was impaired because of our custom unoptimized modifications made to the standard Keras implementation. 

\subsection{Speleothem Dataset}\label{cave_data}

To evaluate RNNs on a real-world generative task, we use the readings of oxygen isotopes obtained from speleothems in the Indian cave \cite{sinha2015trends} dataset. We use the first 1\,800 univariate samples, of which 1\,700 are used for training, 50 for validation, and the last 50 for testing. We trained conventional RNN models for 100 epochs and repeated this process for 10 times with random models' initialization. The optimal number of neurons was found through a grid search to be 30 for the trainable RNN baselines and 50 for the ESN baselines. Grid search was also carried out to find the optimal leaking rate and spectral radius scaling for ESN variants. 

Since ESNs support one-shot learning through linear regression, we exploited the feature that cross-validation (CV) on temporal data can be done very efficiently for them, with minimal overhead \cite{LukoseviciusUselis19,LukoseviciusUselis21}. Concretely, each fold consisted of 50 consecutive samples, and the hyper-parameters were chosen based on the mean validation error. For testing, ESN's readout weights were recomputed from the whole training sequence. %

We have tested three variations of each RNN baseline: regular RNNs, where the model is not aware of any time irregularities, RNNs with $\Delta t_n$ as an additional input, and a variation, where the RNNs received interpolated training sequence as the input. In the interpolated variation (which we refer to as ``Interp.''), the whole training sequence was resampled into regular-time time series with linear interpolation before feeding it in. Then, to evaluate this model, the predicted time series with regular sampling were again resampled with linear interpolation to match the irregular time points of the original time series, and errors were computed.
\setlength{\tabcolsep}{4pt}
\begin{table}[htb]
    \centering
    \caption{Results on the Speleothem dataset }
    \begin{tabular}{l r r r r r r r} 
    \toprule 
     {Model}  &  {Valid. RMSE} &  {Test RMSE} &  Test MAPE & $f(\Delta t_n)$ & Valid. type\\        
        \midrule

ESN    & 0.117 $\pm$ 0.009 & 0.425  $\pm$ 0.064 & 4.311 $\pm$ 0.702 & linear & standard\\
ESN+$\Delta t$    & 0.161 $\pm$ 0.036 & 0.348  $\pm$ 0.044 & 3.377 $\pm$ 0.456 & linear & standard\\
Interp. ESN    & 0.139 $\pm$ 0.022 & 0.346  $\pm$ 0.093 & 3.397 $\pm$ 0.998 & linear & standard\\
\textbf{TSESN}    & 0.129 $\pm$ 0.008 & 0.408  $\pm$ 0.026 & 4.125 $\pm$ 0.296 & exp & standard\\
\textbf{TSESN}    & 0.120 $\pm$ 0.011 & 0.490  $\pm$ 0.021 & 4.952 $\pm$ 0.237 & linear & standard\\
\midrule
GRU & 0.169 $\pm$ 0.010 & 0.351 $\pm$ 0.028 & 3.401 $\pm$ 0.304  & linear & standard \\
LSTM & 0.160 $\pm$ 0.001 & 0.405 $\pm$ 0.009 & 3.999 $\pm$ 0.103  & linear & standard \\
GRU+$\Delta t$ & 0.164 $\pm$ 0.003 & 0.366 $\pm$ 0.017 & 3.559 $\pm$ 0.192  & linear & standard \\
LSTM+$\Delta t$ & 0.162 $\pm$ 0.001 & 0.403 $\pm$ 0.023 & 3.981 $\pm$ 0.267  & linear & standard \\
Interp. GRU & 0.171 $\pm$ 0.010 & \underline{0.334 $\pm$ 0.022} & \underline{3.206 $\pm$ 0.236}  & linear & standard \\
Interp. LSTM & 0.157 $\pm$ 0.002 & 0.383 $\pm$ 0.010 & 3.738 $\pm$ 0.118  & linear & standard \\
\textbf{TSGRU} & 0.187 $\pm$ 0.021 & 0.349 $\pm$ 0.053 & 3.389 $\pm$ 0.577  & linear & standard \\
\midrule
ESN    & 0.193 $\pm$ 0.005 & 0.185  $\pm$ 0.006 & 1.709 $\pm$ 0.057 & exp & CV\\
ESN+$\Delta t$    & 0.201 $\pm$ 0.005 & 0.244  $\pm$ 0.001 & 2.340 $\pm$ 0.010 & exp & CV\\
Interp. ESN    & 0.183 $\pm$ 0.004 & 0.304  $\pm$ 0.005 & 2.917 $\pm$ 0.051 & exp & CV\\
\textbf{TSESN}    & 0.199 $\pm$ 0.008 &\underline{0.159 $\pm$ 0.002}& \underline{1.478 $\pm$ 0.017} & exp & CV\\
\textbf{TSESN}    & 0.182 $\pm$ 0.004 & 0.346  $\pm$ 0.014 & 3.333 $\pm$ 0.150 & linear & CV\\
\bottomrule
    \end{tabular}
    \label{tab:res_copernicus}        
\end{table}

From the results in Table \ref{tab:res_copernicus} it can be seen that among the standard-validated methods, the interpolated GRU model yielded the best results on the testing set, with interpolated ESN, ESN+$\Delta t$, and TSGRU not far behind. 

However, ESN-based methods' ability to be efficiently cross-validated \cite{LukoseviciusUselis19,LukoseviciusUselis21} gave the best effect here. This probably has to do with (the lack of) stationarity of this eons-long time series. Cross-validated TSESN with exponential time function produced by far the best testing results among all the tried methods.

\section{Discussion}\label{dicussion}

Our analytically derived models of task-synchronized RNNs tested with success in thorough numerical simulations with both synthetic and real-world non-uniform time data.

The same task-synchronized RNN idea is in principle applicable to other RNN architectures like BPTT-trained \cite{Werbos90} RNN or LSTM. But in this work, we preferred ESN over fully-trained RNN, because it is simpler and efficient to train and cross-validate, and GRU over LSTM, because it is simpler and closer to ESN. Task-synchronizing other RNN models is a promising future work.

Our models also have some limitations. 
    
As mentioned, large time gaps are problematic. This is mitigated by the nonlinear transformation of $\Delta t_n$ introduced in Section \ref{fdt}, but in some applications filling these big gaps in data by some type of imputation might be the way to go. An analogy could be building a bridge: the span ($\Delta t_n$) between two subsequent piles (data points) can vary to an extent, but if it is too large, we may need to build an additional pile (impute a data point) or several in the middle. 
    
The asynchronous arrival of data in different input dimensions is not handled by this class of models without interpolating/infilling the data.
    
$\Delta t_n$ variations in the data are potentially invisible to our models, which is often an advantage, but can in some cases have useful information in themselves. %

The time-discretization of the models could also potentially be done using a more sophisticated differential equation solver like in \cite{NEURIPS2020_4a5876b4} or \cite{Schake22mthesis} than the straightforward linear Euler approximation used here. 

The advantages of the current approach, however, are its speed, directness, and simplicity. It allows dealing with time-irregular data systematically and effectively while training and using different types of task-synchronized RNNs in virtually the same way as the classical discrete-time ones, and not inventing (interpolating/infilling) new data.

We share the source code at \\ \url{https://github.com/oshapio/task-synchronized-RNNs}.

\bibliographystyle{splncs04}
\bibliography{ML2}

\begin{thebibliography}{10}
\providecommand{\url}[1]{\texttt{#1}}
\providecommand{\urlprefix}{URL }
\providecommand{\doi}[1]{https://doi.org/#1}

\bibitem{BaytasEtAl17}
Baytas, I.M., Xiao, C., Zhang, X., Wang, F., Jain, A.K., Zhou, J.: Patient subtyping via time-aware {LSTM} networks. In: Proceedings of the 23rd ACM SIGKDD International Conference on Knowledge Discovery and Data Mining. pp. 65--74. ACM (2017)

\bibitem{che2018recurrent}
Che, Z., Purushotham, S., Cho, K., Sontag, D., Liu, Y.: Recurrent neural networks for multivariate time series with missing values. Scientific reports  \textbf{8}(1), ~6085 (2018)

\bibitem{chen2018neural}
Chen, T.Q., Rubanova, Y., Bettencourt, J., Duvenaud, D.K.: Neural ordinary differential equations. In: Advances in Neural Information Processing Systems. pp. 6571--6583 (2018)

\bibitem{ChoEtal14}
Cho, K., van Merrienboer, B., G{\"{u}}l{\c{c}}ehre, {\c{C}}., Bougares, F., Schwenk, H., Bengio, Y.: Learning phrase representations using {RNN} encoder-decoder for statistical machine translation. CoRR  \textbf{abs/1406.1078} (2014), \url{http://arxiv.org/abs/1406.1078}

\bibitem{cinar2018period}
Cinar, Y.G., Mirisaee, H., Goswami, P., Gaussier, E., A{\"\i}t-Bachir, A.: Period-aware content attention {RNN}s for time series forecasting with missing values. Neurocomputing  \textbf{312},  177--186 (2018)

\bibitem{FUNAHASHI1993801}
ichi Funahashi, K., Nakamura, Y.: Approximation of dynamical systems by continuous time recurrent neural networks. Neural Networks  \textbf{6}(6),  801--806 (1993). \doi{https://doi.org/10.1016/S0893-6080(05)80125-X}, \url{https://www.sciencedirect.com/science/article/pii/S089360800580125X}

\bibitem{futoma2017learning}
Futoma, J., Hariharan, S., Heller, K.: Learning to detect sepsis with a multitask {G}aussian process {RNN} classifier. In: Proceedings of the 34th International Conference on Machine Learning-Volume 70. pp. 1174--1182. JMLR. org (2017)

\bibitem{Futoma2017}
Futoma, J., Hariharan, S., Sendak, M., Brajer, N., Clement, M., Bedoya, A., O'Brien, C., Heller, K.: An improved multi-output {G}aussian process {RNN} with real-time validation for early sepsis detection. arXiv preprint arXiv:1708.05894  (2017)

\bibitem{GodfreyGashler18}
Godfrey, L.B., Gashler, M.S.: Neural decomposition of time-series data for effective generalization. IEEE transactions on neural networks and learning systems  \textbf{29}(7),  2973--2985 (2018)

\bibitem{Jaeger01a}
Jaeger, H.: The ``echo state'' approach to analysing and training recurrent neural networks. Tech. Rep. GMD Report 148, German National Research Center for Information Technology (2001), \url{https://www.ai.rug.nl/minds/uploads/EchoStatesTechRep.pdf}

\bibitem{JaegerHaas04}
Jaeger, H., Haas, H.: Harnessing nonlinearity: predicting chaotic systems and saving energy in wireless communication. Science  \textbf{304}(5667),  78--80 (2004). \doi{10.1126/science.1091277}, \url{https://www.ai.rug.nl/minds/uploads/ESNScience04.pdf}

\bibitem{JaegerEtAl07si}
Jaeger, H., Luko\v{s}evi\v{c}ius, M., Popovici, D., Siewert, U.: Optimization and applications of echo state networks with leaky-integrator neurons. Neural Networks  \textbf{20}(3),  335--352 (2007)

\bibitem{Kag_2021_CVPR}
Kag, A., Saligrama, V.: Time adaptive recurrent neural network. In: Proceedings of the IEEE/CVF Conference on Computer Vision and Pattern Recognition (CVPR). pp. 15149--15158 (June 2021)

\bibitem{NEURIPS2020_4a5876b4}
Kidger, P., Morrill, J., Foster, J., Lyons, T.: Neural controlled differential equations for irregular time series. In: Larochelle, H., Ranzato, M., Hadsell, R., Balcan, M., Lin, H. (eds.) Advances in Neural Information Processing Systems. vol.~33, pp. 6696--6707. Curran Associates, Inc. (2020), \url{https://proceedings.neurips.cc/paper_files/paper/2020/file/4a5876b450b45371f6cfe5047ac8cd45-Paper.pdf}

\bibitem{lipton2015learning}
Lipton, Z.C., Kale, D.C., Elkan, C., Wetzel, R.: Learning to diagnose with {LSTM} recurrent neural networks. arXiv preprint arXiv:1511.03677  (2015)

\bibitem{liu2009uwave}
Liu, J., Zhong, L., Wickramasuriya, J., Vasudevan, V.: uwave: Accelerometer-based personalized gesture recognition and its applications. Pervasive and Mobile Computing  \textbf{5}(6),  657--675 (2009)

\bibitem{Lorenz63}
Lorenz, E.N.: Deterministic nonperiodic flow. Journal of Atmospheric Science  \textbf{20},  130--141 (1963)

\bibitem{Lukosevicius12a}
Luko{\v{s}}evi{\v{c}}ius, M.: A practical guide to applying echo state networks. In: Montavon, G., Orr, G.B., M{\"{u}}ller, K.R. (eds.) Neural Networks: Tricks of the Trade, 2nd Edition, LNCS, vol.~7700, pp. 659--686. Springer (2012). \doi{10.1007/978-3-642-35289-8_36}, \url{http://dx.doi.org/10.1007/978-3-642-35289-8_36}

\bibitem{LukoseviciusEtal06}
Luko{\v{s}}evi{\v{c}}ius, M., Popovici, D., Jaeger, H., Siewert, U.: Time warping invariant echo state networks. Tech. Rep. No. 2, Jacobs University Bremen (May 2006), \url{https://www.ai.rug.nl/minds/uploads/techreport2.pdf}

\bibitem{LukoseviciusUselis19}
Luko\v{s}evi\v{c}ius, M., Uselis, A.: Efficient cross-validation of echo state networks. In: Artificial Neural Networks and Machine Learning \textendash ICANN 2019: Workshop and Special Sessions. ICANN 2019. Lecture Notes in Computer Science, vol. 11731, pp. 121--133. Springer, Cham (2019). \doi{10.1007/978-3-030-30493-5_12}, \url{https://link.springer.com/chapter/10.1007/978-3-030-30493-5_12}, presentation slides included

\bibitem{LukoseviciusUselis21}
Luko\v{s}evi\v{c}ius, M., Uselis, A.: Efficient implementations of echo state network cross-validation. Cognitive Computation pp. 1--15 (2021). \doi{10.1007/s12559-021-09849-2}, \url{https://link.springer.com/article/10.1007/s12559-021-09849-2}

\bibitem{MackeyGlass77}
Mackey, M.C., Glass, L.: Oscillation and chaos in physiological control systems. Science  \textbf{197}(4300),  287--289 (1977)

\bibitem{mozer2017discrete}
Mozer, M.C., Kazakov, D., Lindsey, R.V.: Discrete event, continuous time {RNN}s. arXiv preprint arXiv:1710.04110  (2017)

\bibitem{Neil:2016:PLA:3157382.3157532}
Neil, D., Pfeiffer, M., Liu, S.C.: Phased {LSTM}: Accelerating recurrent network training for long or event-based sequences. In: Proceedings of the 30th International Conference on Neural Information Processing Systems. pp. 3889--3897. NIPS'16, Curran Associates Inc., USA (2016), \url{http://dl.acm.org/citation.cfm?id=3157382.3157532}

\bibitem{DBLP:journals/corr/PhamTPV16}
Pham, T., Tran, T., Phung, D.Q., Venkatesh, S.: Deepcare: {A} deep dynamic memory model for predictive medicine. CoRR  \textbf{abs/1602.00357} (2016)

\bibitem{NEURIPS2019_42a6845a}
Rubanova, Y., Chen, R.T.Q., Duvenaud, D.K.: Latent ordinary differential equations for irregularly-sampled time series. In: Wallach, H., Larochelle, H., Beygelzimer, A., d\textquotesingle Alch\'{e}-Buc, F., Fox, E., Garnett, R. (eds.) Advances in Neural Information Processing Systems. vol.~32. Curran Associates, Inc. (2019), \url{https://proceedings.neurips.cc/paper_files/paper/2019/file/42a6845a557bef704ad8ac9cb4461d43-Paper.pdf}

\bibitem{Schake22mthesis}
Schake, E.J.: Recurrent neural network approaches to irregularly sampled data. Master's thesis, Kaunas University of Technology, Kaunas, Lithuania (2022), \url{https://epubl.ktu.edu/object/elaba:132820469/}

\bibitem{pmlr-v162-schirmer22a}
Schirmer, M., Eltayeb, M., Lessmann, S., Rudolph, M.: Modeling irregular time series with continuous recurrent units. In: Chaudhuri, K., Jegelka, S., Song, L., Szepesvari, C., Niu, G., Sabato, S. (eds.) Proceedings of the 39th International Conference on Machine Learning. Proceedings of Machine Learning Research, vol.~162, pp. 19388--19405. PMLR (17--23 Jul 2022), \url{https://proceedings.mlr.press/v162/schirmer22a.html}

\bibitem{shukla2018interpolation}
Shukla, S.N., Marlin, B.: Interpolation-prediction networks for irregularly sampled time series. In: International Conference on Learning Representations (2019), \url{https://openreview.net/forum?id=r1efr3C9Ym}

\bibitem{sinha2015trends}
Sinha, A., Kathayat, G., Cheng, H., Breitenbach, S.F., Berkelhammer, M., Mudelsee, M., Biswas, J., Edwards, R.L.: Trends and oscillations in the indian summer monsoon rainfall over the last two millennia. Nature communications  \textbf{6}, ~6309 (2015)

\bibitem{TallecOllivier18}
Tallec, C., Ollivier, Y.: Can recurrent neural networks warp time? arXiv preprint arXiv:1804.11188  (2018)

\bibitem{WEERAKODY2021161}
Weerakody, P.B., Wong, K.W., Wang, G., Ela, W.: A review of irregular time series data handling with gated recurrent neural networks. Neurocomputing  \textbf{441},  161--178 (2021). \doi{10.1016/j.neucom.2021.02.046}, \url{https://www.sciencedirect.com/science/article/pii/S0925231221003003}

\bibitem{Werbos90}
Werbos, P.J.: Backpropagation through time: what it does and how to do it. Proceedings of the IEEE  \textbf{78}(10),  1550--1560 (1990)

\bibitem{NIPS1995_7cce53cf}
Williams, C., Rasmussen, C.: Gaussian processes for regression. In: Touretzky, D., Mozer, M., Hasselmo, M. (eds.) Advances in Neural Information Processing Systems. vol.~8. MIT Press (1995), \url{https://proceedings.neurips.cc/paper_files/paper/1995/file/7cce53cf90577442771720a370c3c723-Paper.pdf}

\bibitem{ijcai2017-504}
Zhu, Y., Li, H., Liao, Y., Wang, B., Guan, Z., Liu, H., Cai, D.: What to do next: Modeling user behaviors by time-{LSTM}. In: Proceedings of the Twenty-Sixth International Joint Conference on Artificial Intelligence, {IJCAI-17}. pp. 3602--3608 (2017). \doi{10.24963/ijcai.2017/504}, \url{https://doi.org/10.24963/ijcai.2017/504}

\end{thebibliography}

\end{document}